\documentclass{article}
\usepackage{spconf,amsmath}

\usepackage{amssymb}
\usepackage{latexsym}
\usepackage{graphicx}
\graphicspath{{images/}}

\usepackage{url}
\usepackage{xcolor}
\definecolor{newcolor}{rgb}{.8,.349,.1}

\usepackage{array}
\newcolumntype{P}[1]{>{\centering\arraybackslash}p{#1}}
\newcolumntype{M}[1]{>{\centering\arraybackslash}m{#1}}
\def\rot#1{\rotatebox{90}{#1}}

\def\x{{\mathbf x}}

\def\x{{\mathbf x}}
\def\y{{\mathbf y}}
\def\z{{\mathbf z}}
\def\A{{\mathbf A}}
\def\n{{\mathbf n}}

\title{Regularization via deep generative models: an analysis point of view}
\name{Thomas Oberlin and Mathieu Verm\thanks{Part of this work has been funded by the Institute for Artificial and Natural Intelligence Toulouse (ANITI) under grant agreement ANR-19-PI3A-0004.
}}
\address{ISAE-SUPAERO, Universit\'e de Toulouse, 10 Avenue Edouard Belin, Toulouse 31400, France}
\begin{document}
%\ninept
%
%\copyrightnotice

\maketitle
\begin{abstract}
This paper proposes a new way of regularizing an inverse problem in imaging (e.g., deblurring or inpainting) by means of a deep generative neural networks. Compared to end-to-end models, such approaches seem particularly interesting since the same network can be used for \emph{many} different problem and experimental conditions, as soon as the generative model is suited to the data.
Previous works proposed to use a \emph{synthesis} framework, where the estimation is performed on the latent vector, the solution being obtained afterwards via the decoder. Instead, we propose an \emph{analysis} formulation where we directly optimize the image itself and penalize the latent vector.
We illustrate the interest of such a formulation by running experiments of inpainting, deblurring and super-resolution. In many cases our technique achieves a clear improvement of the performance and seems to be more robust, in particular with respect to initialization.
\end{abstract}

\section{Introduction}
\label{sec1}

Inverse problems are ubiquitous in imaging, because many image acquisition pipelines involve degradation operators that need to be inverted afterwards, such as radial projections in tomography or spatial invariant blur in any image or signal acquisition sensor \cite{bertero2020introduction}. This task can be seen as an estimation problem, which is often solved via optimization algorithms or Monte Carlo sampling \cite{idier2013bayesian}. To circumvent the ill-posedness of the problem, practicioners need \emph{priors} or \emph{regularizers} that will promote some behavior of the solution.  
    
Famous generic priors have been proposed in imaging in the last decades, such as Total Variation \cite{rudin1992nonlinear} and variants \cite{bredies2010total}, which favour piecewise-smooth images. This is done by minimizing the $\ell_1$ norm of some linear differentiation operator of the image. If we know a priori a basis in which the data is sparse, the $\ell_1$ norm can also be placed on the representation coefficients in that basis, which remains an efficient technique in some applications \cite{monier2020fast}. More recently, significant improvements have been achieved by considering  \emph{plug-and-play priors} and variants \cite{venkatakrishnan2013plug,chan2016plug}, often based on highly efficient denoisers such as BM3D \cite{dabov2007image}.
    
But in many situations such a basis is not known, although it can be \emph{learned} from some available training data. This idea has been first introduced with the problem of dictionary learning \cite{aharon2006k}, which simultaneously learns the dictionary and the noise-free image. It has then been applied to many inverse problems by training the dictionary beforehands on a representative dataset. More recently, numerous works considered to train a deep neural networks (DNN) instead of a dictionary, which appeared to be a more efficient and effective model for images. Some works proposed to learn a denoising networks \cite{zhang2017learning,zhang2018ffdnet} and use it as a plug-and-play prior; or to learn a 1-class classification DNN \cite{rick2017one} which is used as a projection operator onto the set of natural images. 
    
We will focus on a different approach introduced by \cite{bora2017compressed}, which simply learns a generative DNN and use it to constrain the solution to live in its range. In this work, authors showed recovery results similar to the ones obtained in compressed sensing, and demonstrated the effectiveness of the approach with a variational autoencoder (VAE) and a Deep Convolutional Generative Adversarial Network (DCGAN). Impressive results have been obtained by considering another kind of generative model termed Glow \cite{asim2019invertible}, which is invertible and thus enables to better control the distribution of the latent factor by optimizing the maximum likelihood. Our aim is to show that with such a network, the result can be dramatically improved by optimizing the image itself instead of the latent code, the network being used only for regularization purpose. We name our approach deep \emph{analysis} regularization, as opposed to the \emph{synthesis} regularization proposed in \cite{bora2017compressed}, and following the formulation used in sparse recovery \cite{elad2007analysis,nam2013cosparse}.

Note that apart from the "deep plug-and-play approaches" discussed above, which use deep learning only in the regularization, there are several lines of works that tackle inverse problems with deep learning in very different ways. To cite a few, this includes unrolled deep networks \cite{bertocchi2020deep}, unsupervised approaches \cite{ulyanov2018deep} or other end-to-end approaches \cite{yu2018generative}. Each line of methods has its pros and cons, and comparing those techniques is beyond the scope of the current paper. In a nutshell, plug-and-play approaches seem more generic (one prior for different problems) and somehow more grounded (we can control the optimization algorithm), but they might be outperformed by problem-dependant methods \cite{arridge2019solving}.

The remainder of this paper is structured as follows. Section \ref{sec2} recalls some background and the works upon which we build our analysis deep regularization technique, presented in Section \ref{sec3}. The experimental setting and the corresponding results are then presented in Sections \ref{sec4} and \ref{sec5}, respectively, while a summary and some perspectives are discussed in Section \ref{sec6} which concludes the paper.

\section{Background}
\label{sec2}

\subsection{Problem statement}

Many imaging problems consists in sensing the image $\y$ from an underlying true scene $\x$ according to
\begin{equation}
    \label{eq:model}
    \y = \A \x + \n ,
\end{equation}
where $\A$ is some observation operator which is often known with good accuracy, and $\n$ an error term accounting for random effects and potential model mismatch. This model encompasses denoising (A is the identity matrix), inpainting (A is a mask), deblurring (A is Toeplitz), etc. The noise term is often assumed to be independent and Gaussian, although in many imaging modalities is contains a mixing of Gaussian and Poisson components \cite{makitalo2012optimal}.

The inverse problem consists in estimating the true scene $\x$, which amounts to invert the linear operator $\A$, which is in general singular or at least badly conditioned. To overcome this ill-posedness, one needs to bring additional information by means of a \emph{regularizer} $\varphi$. The formulation of the inverse problem then writes
\begin{equation}
    \label{eq:ip}
    \hat \x  =\arg \min_\x \frac{1}{2}\left\|\A\x-\y \right\|_2^2 + \lambda \varphi(\x)\,
\end{equation}
where parameter $\lambda$ tunes the level of regularization. 

Such a formulation can be related to Bayesian estimation: if the noise $\n$ is assumed Gaussian, then (\ref{eq:ip}) is the maximum a posteriori (MAP) estimate under the prior $x\sim e^{-\varphi}$. Note that this remain valid if the noise follows a different statistics, as long as we replace the least-square by the adequate divergence measure. For convenience, we will restrict our study to the least-square formulation, which is the most convenient and the most widely used.

\subsection{Priors via deep generative neural networks}

Auto-encoders are special kinds of neural networks. They are composed of an encoder which computes the latent code $\z = E(\x)$, and a decoder or a generator $D(\z)$, both being trained so that $\x \approx D(E(\x))$ for all data samples $\x$ of a given training dataset.
While autoencoders have been introduced a long time ago, there has been a stunning renewal with the introduction of so-called variational auto-encoders (VAEs) in \cite{kingma2013auto,rezende2014stochastic}. The aim of these works was to constrain the latent space to exhibit a proper structure, which can be seen as a regularization of the learning problem. The first purpose was to be able to generate realistic images from random vectors, which requires to control the distribution of the latent codes. Recall that the loss function of a VAE is intractable and requires approximate inference techniques. To overcome this, some works \cite{kingma2016improved,kingma2018glow} followed a different approach inspired by normalizing flows, where each layer of the network remains invertible, its gradient being efficiently computed for backpropagation. Once learned, such networks can be use as priors to regularize any inverse problem, as described in the following Section.

\subsection{Synthesis-based regularization}
We describe here our baselines \cite{bora2017compressed,asim2019invertible}. They both assume that a generative model has been learned beforehands on a representative database. Then, they propose to solve (\ref{eq:ip}), by constraining the sought image $\x$ to be in the range of the generator: $\x=D(\z)$ for some latent vector $\z$. Both works add an extra regularizer, based on the normal distribution assumed for the latent vector, which finally gives
\begin{equation}
    \label{eq:synth}
    \hat \x  = D \left( \arg \min_\z \frac{1}{2}\left\|\A D(\z)-\y \right\|_2^2 + \lambda \|\z\|_2^2 \right).
\end{equation}

\section{An analysis formulation}
\label{sec3}

\subsection{Limits of the synthesis regularization}
The synthesis formulation (\ref{eq:synth}) operates in the latent space, which brings some clear benefits but also several drawbacks. On the one hand, the solution of such a problem is likely to be visually consistent, since it has been generated by the decoder. But on the other hand, the found solution is often very different from the ground truth, because of the strong non-convexity of the problem which makes the solution highly dependent to initialization. To reduce this effect one should properly initialize the optimization algorithm, for instance by setting $\z_0 = E(\x_0)$, with $x_0$ an initial guess. According to our experience, this trick however does not always produce an image which is fully consistent with the observations.

\subsection{An analysis formulation}
To circumvent the drawbacks discussed above, we propose here a straightforward solution which consists in optimizing directly the image. The regularization is performed in the latent space, capitalizing on the Gaussian distribution which is assumed for the latent vector. This writes:
\begin{equation}
    \label{eq:anal}
    \hat \x  =  \arg \min_\x \frac{1}{2}\left\|\A \x-\y \right\|_2^2 + \lambda \|E(\x)\|_2^2.
\end{equation}
We named this formulation \emph{analysis}. Note that, contrary to the synthesis formulation, the role of parameter $\lambda$ is fundamental, since it controls the only regularization brought about by the network.

The two formulations are the counterpart of what was already studied concerning sparse representations in redundant dictionaries \cite{elad2007analysis}. But in the case of deep regularization, both approaches always lead to different results, even if the network is invertible, because of the strong non-convexity of the problem.
The remaining of the letter is devoted to illustrate the pros and cons of the two formulations, on various simple inverse problems. To this end, we describe the experimental setting and the results in the next two sections.

\section{Experimental setting}
\label{sec4}

We restrict our study to the comparison between the proposed analysis formulation and the more standard synthesis, as implemented by \cite{asim2019invertible}. Note that this recent work competes favorably with several state of the art techniques, we thus tried to reproduce a close experimental setting.

For the generative DNN, we use the Glow network as introduced in \cite{kingma2018glow}. Glow is inspired by flow-based generative models \cite{dinh2014nice}, whose particularity is to exhibit a tractable log-likelihood for the generative process, which avoids the use of approximate inference techniques. This is achieved by means of a particular architecture, composed of invertible layers with (fast) tractable Jacobian. Instead of standard convolutions, the layers of Glow are thus composed of split, $1\times 1$ convolutions and actnorm steps, we refer to \cite{kingma2018glow} for more details.
We trained Glow on the CelebA dataset \cite{CelebA}, composed of colored images of faces with size $64\times64$. We used 4 blocks with 32 steps of flow each and additive coupling layers. The optimizer chosen for the training was Adam with a learning rate of $10^{-4}$ ($\beta_1 = 0.9$ \& $\beta_2 = 0.999$). 

Insofar as Glow is invertible, it can be used both as an Encoder $E(.)$ or a Decoder $D(.)$ for (respectively) the analysis and the synthesis formulations. Because of the Gaussian prior on the latent space, the \textit{regularizer} $\varphi$ will be the squared $\ell_2$ norm $\varphi(x)= \|\x\|^2_2$. The optimisation problems (\ref{eq:synth}) and (\ref{eq:anal}) are solved by gradient descent (either on the latent space or on the image). The initialization of the gradient descent is done, as explained by \cite{asim2019invertible}, by $\z^0=0$ and $\x^0=D(0)$ for the synthesis and analysis formulations, respectively. Parameter $\lambda$ is chosen empirically around $10^{-4}$ for each method and every inverse problems, so as to obtain the best performance.

In the experiments, we consider the three following inverse problems:
        \begin{itemize}
            \item Super-resolution: increasing the resolution of an image which has been previously downsampled by a factor 2 or 4 with local averaging (uniform filter).
            \item Deblurring: removing the blur caused by a $7\times7$ uniform filter.
            \item Inpainting: filling the gaps in an image caused by the application of a mask. We consider two different settings, either 60\% of uniformly random missing pixels or a squared mask centered on the image of size $\frac{1}{9} \times \frac{1}{9}$.
        \end{itemize}
        
The metrics used to evaluate the effectiveness of the result are the widely used PSNR and SSIM. For the sake of reproducibility, we will release our code as soon as the paper is accepted.

\section{Results}
\label{sec5}

\begin{figure}[h]
\centering \scriptsize
\rot{\qquad True} \includegraphics{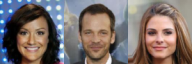} \, \includegraphics{original.png}\\
\rot{\quad  Observed} \includegraphics{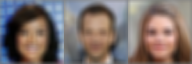} \, \includegraphics{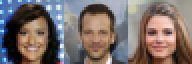}\\
\rot{\quad Synthesis} \includegraphics{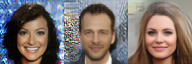}\,  \includegraphics{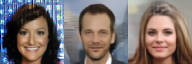}\\
\rot{\quad Analysis} \includegraphics{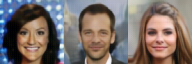} \, \includegraphics{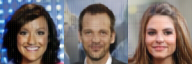}
\caption{Results of deblurring (left) and 2x super-resolution (right) for three images of the test set. From top to bottom: truth, degraded image, results of synthesis and analysis formulations.}
\label{fig:deb}
\end{figure}

Let us first show visual comparisons between the synthesis and analysis frameworks, to highlight the pros and cons of both formulations. The result of the deblurring and the super-resolution experiments are depicted on Figure \ref{fig:deb} for three images of the test set, that were randomly selected. 
Concerning the deblurring, the results of the synthesis method exhibit artefacts, that are particularly visible in the background and do not disappear if we increase $\lambda$. The analysis formulation instead shows a far better result, both for the background and the face. To explain this, we believe that the synthesis approach struggles to represent the background, in particular for the first image because its pattern is a bit unusual and might not have been seen during training. In the analysis approach instead there is no need to generate the background, since the optimization operated within the image space. 
The super-resolution experiment investigates a more difficult setting where fewer information is available, and for this the synthesis approach might seem preferable. Indeed, the faces reconstructed by synthesis are much more pleasant than the analysis counterpart, which suffer from strong artefacts (the prior does not seem to work well enough, even for a large value of $\lambda$). Note however that the synthesis results, although visually better, are outperformed by analysis in terms of PSNR or SSIM, as we will see later. 

\begin{table*}
\centering
\begin{tabular}{l|c|c|c|c}
\hline Task & PSNR (synthesis) & PSNR (analysis) & SSIM (synthesis) & SSIM (analysis) \\ \hline
Deblurring & 23.39 {\small $\pm 2.11$} & \textbf{32.13} {\small $\pm 1.95$} & 0.74 {\small $\pm 0.11$} & \textbf{0.94} {\small $\pm 0.01$}\\
Super-resolution (x2) &20.85 {\small $\pm 6.45$} & \textbf{31.20} {\small $\pm 1.60$} & 0.68 {\small $\pm 0.26$} & \textbf{0.93} {\small $\pm 0.01$}\\
Super-resolution (x4) & 20.74 {\small $\pm 3.46$} & \textbf{24.19} {\small $\pm 1.22$} & 0.67 {\small $\pm 0.11$} & \textbf{0.75} {\small $\pm 0.02$}\\ 
Inpainting (random mask) & 22.29 {\small $\pm 3.59$} & \textbf{27.64} {\small $\pm 2.48$} & 0.74 {\small $\pm 0.13$} & \textbf{0.88} {\small $\pm 0.04$} \\ 
Inpainting (structured mask) & \textbf{31.49} {\small $\pm 2.50$} & 27.94 {\small $\pm 3.25$} & \textbf{0.95} {\small $\pm 0.02$} & 0.92 {\small $\pm 0.03$}\\ \hline
\end{tabular}
\caption{Average performance over a test set of 15 images, along with the standard deviation. The best score between analysis and synthesis are highlighted in bold, for both metrics.}
\label{tab}
\end{table*}
    
\begin{figure}[h]
\centering
\scriptsize
\rot{\qquad True} \includegraphics{original.png} \, \includegraphics{original.png} \\
\rot{\quad Observed} \includegraphics{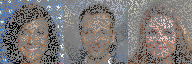} \, \includegraphics{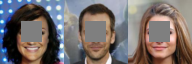} \\
\rot{\quad Synthesis} \includegraphics{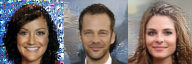} \, \includegraphics{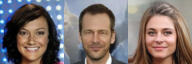} \\
\rot{\quad Analysis} \includegraphics{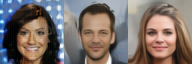} \, \includegraphics{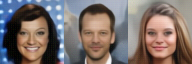}
\caption{Inpainting with random mask (left) and structured mask (right), for the same three images of the test set.}
\label{fig:in1}
\end{figure}

Let us now move to the inpainting results, depicted on Figure \ref{fig:in1} for the random and the structured masks. In the first case, the analysis technique seems to achieve the best results, while it is outperformed by its synthesis counterpart for the structured mask. We have the same explanations as in the previous experiment: we believe that in general, the analysis formulation stays closer to the true solution because it does not suffer from the intrinsic bias given by the generative network, and its optimization seems safer (there might be more spurious local minima in the latent space than in the image space). However when the problem becomes too difficult, the analysis does not seem to regularize enough, and the synthesis formulation obtains visually better results, although not clearly better in terms of PSNR or SSIM.

We also computed the PSNR and SSIM for 15 different images and the 5 different inverse problems, and gather the results in Table \ref{tab}. It confirms the previous findings: in most cases, the analysis formulation enables an image recovery which is dramatically closer to the ground truth than with the synthesis formulation. But for the most difficult problem, i.e., inpainting with a structured mask (and to a lower extent, super-resolution with factor 4), the synthesis formulation seems better. Another interesting point is the variance of the metrics across the 15 images, which is significantly lower for the analysis formulation. This means that the optimization landscape seems somehow safer in the image space that in the latent space, which is an expected benefit of the analysis formulation.

\section{Conclusion}
\label{sec6}

We presented here a variant of the deep learning regularization via generative models, were the solution is searched directly in the image space. According to our experiments, our \emph{analysis} formulation obtains an estimation of the image which is often closer to the ground truth, even if sometimes visually less pleasant, particularly when the observation does not contain enough information. Besides, it seems that the results of the analysis formulation are more robust than with the synthesis framework.

Further work is required to better understand what is the best formulation for a given setting, and to extend these findings to improved learning settings such as \cite{gonzalez2019solving} or \cite{jalal2020robust}. Concerning the applications, the main bottleneck up to now is the limited ability of generative networks to represent heterogoneous datasets, such as images with various size and content. Forthcoming contributions in generative models should soon overcome this, opening a huge number of possible real-world applications.

\bibliographystyle{IEEEbib}
\bibliography{refs}

\begin{thebibliography}{10}

\bibitem{bertero2020introduction}
M.~Bertero,
\newblock {\em Introduction to inverse problems in imaging},
\newblock CRC press, 2020.

\bibitem{idier2013bayesian}
J.~Idier,
\newblock {\em Bayesian approach to inverse problems},
\newblock John Wiley \& Sons, 2013.

\bibitem{rudin1992nonlinear}
L.~I. Rudin, S.~Osher, and E.~Fatemi,
\newblock ``Nonlinear total variation based noise removal algorithms,''
\newblock {\em Physica D: nonlinear phenomena}, vol. 60, no. 1-4, pp. 259--268,
  1992.

\bibitem{bredies2010total}
K.~Bredies, K.~Kunisch, and T.~Pock,
\newblock ``Total generalized variation,''
\newblock {\em SIAM Journal on Imaging Sciences}, vol. 3, no. 3, pp. 492--526,
  2010.

\bibitem{monier2020fast}
E.~Monier, T.~Oberlin, N.~Brun, X.~Li, M.~Tenc{\'e}, and N.~Dobigeon,
\newblock ``Fast reconstruction of atomic-scale {STEM-EELS} images from sparse
  sampling,''
\newblock {\em Ultramicroscopy}, p. 112993, 2020.

\bibitem{venkatakrishnan2013plug}
S.~V. Venkatakrishnan, C.~A. Bouman, and B.~Wohlberg,
\newblock ``Plug-and-play priors for model based reconstruction,''
\newblock in {\em IEEE Global Conference on Signal and Information Processing
  (GlobalSIP)}, 2013, pp. 945--948.

\bibitem{chan2016plug}
S.~H. Chan, X.~Wang, and O.~A. Elgendy,
\newblock ``Plug-and-play {ADMM} for image restoration: Fixed-point convergence
  and applications,''
\newblock {\em IEEE Transactions on Computational Imaging}, vol. 3, no. 1, pp.
  84--98, 2016.

\bibitem{dabov2007image}
K.~Dabov, A.~Foi, V.~Katkovnik, and K.~Egiazarian,
\newblock ``Image denoising by sparse {3-D} transform-domain collaborative
  filtering,''
\newblock {\em IEEE Transactions on Image Processing}, vol. 16, no. 8, pp.
  2080--2095, 2007.

\bibitem{aharon2006k}
M.~Aharon, M.~Elad, and A.~Bruckstein,
\newblock ``{K-SVD}: An algorithm for designing overcomplete dictionaries for
  sparse representation,''
\newblock {\em IEEE Transactions on Signal Processing}, vol. 54, no. 11, pp.
  4311--4322, 2006.

\bibitem{zhang2017learning}
K.~Zhang, W.~Zuo, S.~Gu, and L.~Zhang,
\newblock ``Learning deep {CNN} denoiser prior for image restoration,''
\newblock in {\em IEEE conference on computer vision and pattern recognition
  (CVPR)}, 2017, pp. 3929--3938.

\bibitem{zhang2018ffdnet}
K.~Zhang, W.~Zuo, and L.~Zhang,
\newblock ``{FFDNet}: Toward a fast and flexible solution for {CNN}-based image
  denoising,''
\newblock {\em IEEE Transactions on Image Processing}, vol. 27, no. 9, pp.
  4608--4622, 2018.

\bibitem{rick2017one}
J.H. Rick~Chang, C.-L. Li, B.~Poczos, B.V.K. Vijaya~Kumar, and A.~C.
  Sankaranarayanan,
\newblock ``One network to solve them all--solving linear inverse problems
  using deep projection models,''
\newblock in {\em IEEE International Conference on Computer Vision (CVPR)},
  2017, pp. 5888--5897.

\bibitem{bora2017compressed}
A.~Bora, A.~Jalal, E.~Price, and A.~G. Dimakis,
\newblock ``Compressed sensing using generative models,''
\newblock in {\em International Conference on Machine Learning (ICML)}, 2017,
  pp. 537--546.

\bibitem{asim2019invertible}
M.~Asim, A.~Ahmed, and P.~Hand,
\newblock ``Invertible generative models for inverse problems: mitigating
  representation error and dataset bias,''
\newblock in {\em International Conference on Machine Learning (ICML)}, 2020.

\bibitem{elad2007analysis}
M.~Elad, P.~Milanfar, and R.~Rubinstein,
\newblock ``Analysis versus synthesis in signal priors,''
\newblock {\em Inverse problems}, vol. 23, no. 3, pp. 947, 2007.

\bibitem{nam2013cosparse}
S.~Nam, M.~E. Davies, M.~Elad, and R.~Gribonval,
\newblock ``The cosparse analysis model and algorithms,''
\newblock {\em Applied and Computational Harmonic Analysis}, vol. 34, no. 1,
  pp. 30--56, 2013.

\bibitem{bertocchi2020deep}
C.~Bertocchi, E.~Chouzenoux, M-C. Corbineau, J-C. Pesquet, and M.~Prato,
\newblock ``Deep unfolding of a proximal interior point method for image
  restoration,''
\newblock {\em Inverse Problems}, vol. 36, no. 3.

\bibitem{ulyanov2018deep}
D.~Ulyanov, A.~Vedaldi, and V.~Lempitsky,
\newblock ``Deep image prior,''
\newblock in {\em IEEE Conference on Computer Vision and Pattern Recognition
  (CVPR)}, 2018, pp. 9446--9454.

\bibitem{yu2018generative}
J.~Yu, Z.~Lin, J.~Yang, X.~Shen, X.~Lu, and T.~S. Huang,
\newblock ``Generative image inpainting with contextual attention,''
\newblock in {\em IEEE conference on computer vision and pattern recognition
  (CVPR)}, 2018, pp. 5505--5514.

\bibitem{arridge2019solving}
S.~Arridge, P.~Maass, O.~{\"O}ktem, and C-B. Sch{\"o}nlieb,
\newblock ``Solving inverse problems using data-driven models,''
\newblock {\em Acta Numerica}, vol. 28, pp. 1--174, 2019.

\bibitem{makitalo2012optimal}
M.~Makitalo and A.~Foi,
\newblock ``Optimal inversion of the generalized {A}nscombe transformation for
  {Poisson-Gaussian} noise,''
\newblock {\em IEEE Transactions on Image Processing}, vol. 22, no. 1, pp.
  91--103, 2012.

\bibitem{kingma2013auto}
D.~P. Kingma and M.~Welling,
\newblock ``Auto-encoding variational {B}ayes,''
\newblock in {\em International Conference on Learning Representations (ICLR)},
  2014.

\bibitem{rezende2014stochastic}
D.~J. Rezende, S.~Mohamed, and D.~Wierstra,
\newblock ``Stochastic backpropagation and approximate inference in deep
  generative models,''
\newblock in {\em International Conference on Machine Learning (ICML)}, 2014.

\bibitem{kingma2016improved}
D.~P. Kingma, T.~Salimans, R.~Jozefowicz, X.~Chen, I.~Sutskever, and
  M.~Welling,
\newblock ``Improved variational inference with inverse autoregressive flow,''
\newblock in {\em Advances in neural information processing systems (NeurIPS)},
  2016, pp. 4743--4751.

\bibitem{kingma2018glow}
D.~P. Kingma and P.~Dhariwal,
\newblock ``Glow: Generative flow with invertible 1x1 convolutions,''
\newblock in {\em Advances in neural information processing systems (NeurIPS)},
  2018, pp. 10215--10224.

\bibitem{dinh2014nice}
L.~Dinh, D.~Krueger, and Y.~Bengio,
\newblock ``Nice: Non-linear independent components estimation,''
\newblock {\em arXiv preprint arXiv:1410.8516}, 2014.

\bibitem{CelebA}
Z.~Liu, P.~Luo, X.~Wang, and X.~Tang,
\newblock ``Large-scale celebfaces attributes (celeba) dataset,''
\newblock {\em Retrieved August}, vol. 15, pp. 2018, 2018.

\bibitem{gonzalez2019solving}
M.~Gonzalez, A.~Almansa, M.~Delbracio, P.~Mus{\'e}, and P.~Tan,
\newblock ``Solving inverse problems by joint posterior maximization with a
  {VAE} prior,''
\newblock {\em arXiv preprint arXiv:1911.06379}, 2019.

\bibitem{jalal2020robust}
A.~Jalal, L.~Liu, A.~G. Dimakis, and C.~Caramanis,
\newblock ``Robust compressed sensing of generative models,''
\newblock in {\em Conference on Neural Information Processing Systems
  (NeurIPS)}, 2020.

\end{thebibliography}

\end{document}